\documentclass[11pt,a4paper]{article}

\usepackage[hyperref]{emnlp-ijcnlp-2019}
\usepackage{times}
\usepackage{latexsym}
\usepackage[T1]{fontenc}

\usepackage{url}

\usepackage{amsmath}
\usepackage{multirow}
\usepackage{microtype}
\usepackage{booktabs}
\usepackage[font=small]{caption}
\usepackage{amssymb}
\usepackage{tikz-dependency}
\usepackage{tikz}

\usepackage{xcolor}

\aclfinalcopy 

\newcommand\x{\mathbf{x}}
\newcommand\h{\mathbf{h}}
\newcommand\y{\mathbf{y}}

\title{Designing and Interpreting Probes with Control Tasks}

\author{John Hewitt \\
  Stanford University\\
  {\tt johnhew@stanford.edu} \\\And
  Percy Liang\\
  Stanford University\\
{\tt pliang@cs.stanford.edu} \\}

\date{}

\begin{document}
\maketitle

\begin{abstract}
  Probes, supervised models trained to predict properties (like parts-of-speech) from representations (like ELMo), have achieved high accuracy on a range of linguistic tasks. But does this mean that
  the representations encode linguistic structure or just that the probe has learned the linguistic task?
  In this paper, we propose \textit{control tasks}, which associate word types with random outputs, to complement linguistic tasks. 
  By construction, these tasks can only be learned by the probe itself.
  So a good probe, (one that reflects the representation), should be \emph{selective}, achieving high linguistic task accuracy and low control task accuracy.
  The selectivity of a probe puts linguistic task accuracy in context with the probe's capacity to memorize from word types.
  We construct control tasks for English part-of-speech tagging and dependency edge prediction, and show that popular probes on ELMo representations are not selective.
We also find that dropout, commonly used to control probe complexity, is ineffective for improving selectivity of MLPs,
but that other forms of regularization are effective.
  Finally, we find that while probes on the first layer of ELMo yield slightly better part-of-speech tagging accuracy than the second, probes on the second layer are substantially more selective, which raises the question of which layer better represents parts-of-speech.
  \looseness=-1
\end{abstract}

\begin{figure}
  \centering
  \includegraphics[width=.8\linewidth]{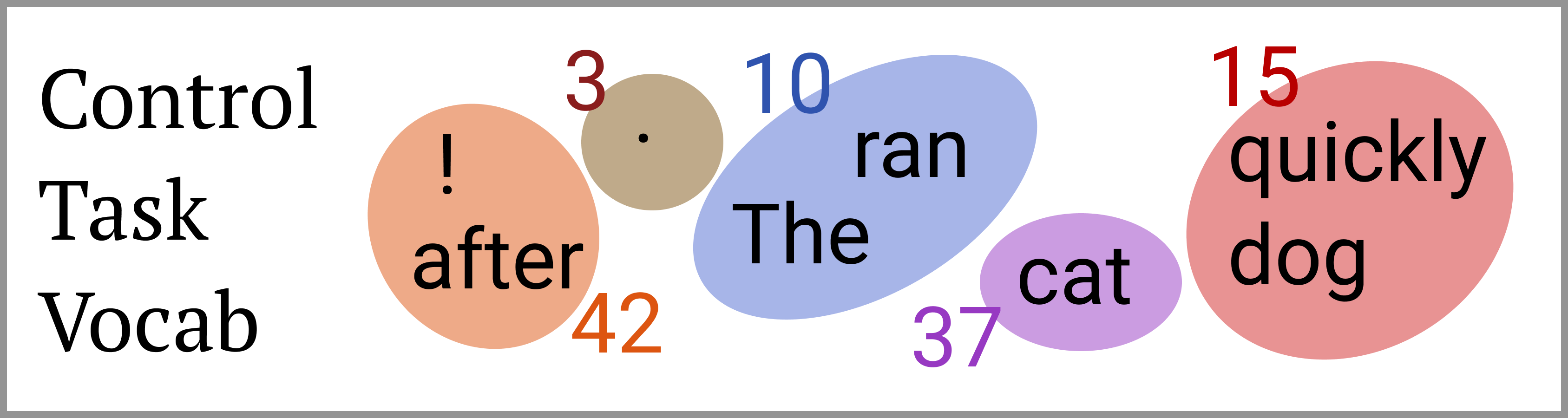}\\
  \vspace{4pt}
  \small
\begin{tabular}{c c c c c c c c c c c c c c c c c}
\toprule
 Sentence 1& The & cat & ran & quickly & .\\
 \bf Part-of-speech & DT & NN & VBD & RB & .\\
 \bf Control task & \color{blue} $\mathbf{10}$ &\color{purple}  $\mathbf{37}$ & \color{blue}$\mathbf{10}$ &\color{red} $\mathbf{15}$ & \color{brown}$\mathbf{3}$ \\
 \midrule
 Sentence 2& The & dog & ran & after & !\\
 \bf Part-of-speech & DT & NN & VBD & IN & .\\
 \bf Control task & \color{blue}$\mathbf{10}$ & \color{red}$\mathbf{15}$ & \color{blue}$\mathbf{10}$ &\color{orange} $\mathbf{42}$ &\color{orange} $\mathbf{42}$ \\
 \bottomrule
\end{tabular}
\caption{\label{figureheaderexplanation} Our control tasks define random behavior (like a random output, top) for each word type in the vocabulary. Each word token is assigned its type's output, regardless of context (middle, bottom.) Control tasks have the same input and output space as a linguistic task (e.g., parts-of-speech) but can only be learned if the probe memorizes the mapping.
}
\end{figure}

\vspace{-10pt}

\section{Introduction}

As large-scale unsupervised representations such as BERT and ELMo improve downstream performance on a wide range of natural language tasks \cite{devlin2019bidirectional,peters2018deep,radford2019language}, what these models learn about language remains an open scientific question.
An emerging body of work investigates this question through \textit{probes}, supervised models trained to predict a property (like parts-of-speech) from a constrained view of the representation.
Probes trained on various representations have obtained high accuracy on tasks requiring part-of-speech and morphological information \cite{belinkov2017what}, syntactic and semantic information \cite{peters2018dissecting,tenney2018what}, among other properties \cite{conneau2018what}, providing evidence that deep representations trained on large datasets are predictive of a broad range of linguistic properties. 
\looseness=-1

But when a probe achieves high accuracy on a linguistic task using a representation, can we conclude that the representation encodes linguistic structure, or has the probe just learned the task?
Probing papers tend to acknowledge this uncertainty, putting accuracies in context using random representation baselines \cite{zhang2018language} and careful task design \cite{hupkes2018visualisation}.
Even so, as long as a representation is a lossless encoding, a sufficiently expressive probe with enough training data can learn \textit{any} task on top of it.

In this paper, we propose \textit{control tasks}, which associate word types with random outputs, to give intuition for the expressivity of probe families and provide insight into how representation and probe interact to achieve high task accuracy.

\begin{figure}
  \centering
  \includegraphics[width=\linewidth]{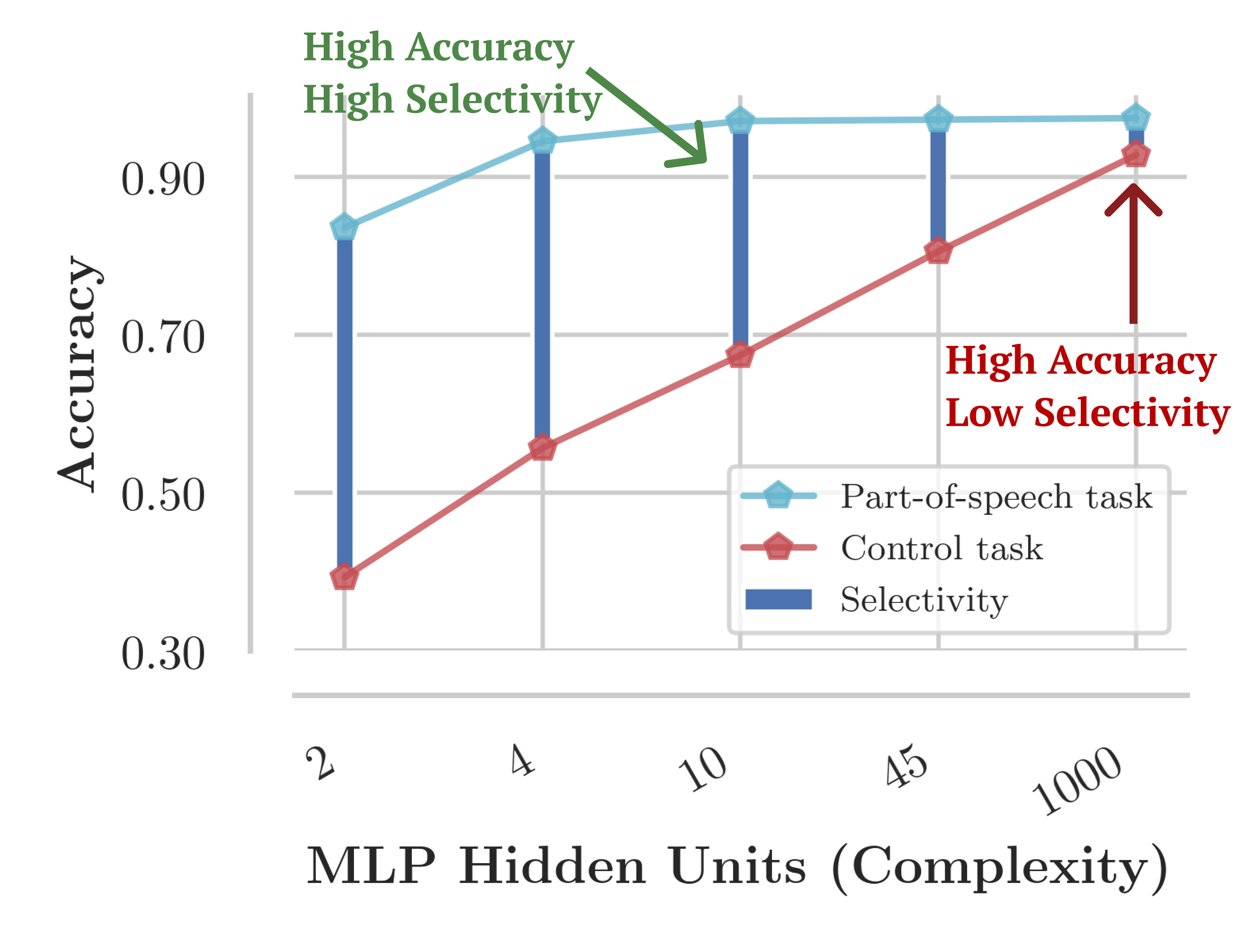}
  \caption{\label{figureselectivity}
    Selectivity is defined as the difference between linguistic task accuracy and control task accuracy, and can vary widely, as shown, across probes which achieve similar linguistic task accuracies.
  These results taken from \S~\ref{sectionselectivityresults}.}
\end{figure}

Control tasks are based on the intuition that the more a probe is able to make task output decisions independently of the linguistic properties of a representation, the less its accuracy on a linguistic task necessarily reflects the properties of the representation.
Thus, a good probe (one that provides insights into the linguistic properties of a representation) should be what we call \textit{selective}, achieving high linguistic task accuracy and low control task accuracy (see Figure~\ref{figureselectivity}).

We show that selectivity can be a guide in designing probes and interpreting probing results, complementary to random representation baselines; as of now, there is little consensus on how to design probes.
Early probing papers used linear functions \cite{shi2016does,ettinger2016probing,alain2016understanding}, which are still used \cite{bisazza2018lazy,liu2019linguistic}, but multi-layer perceptron (MLP) probes are at least as popular \cite{belinkov2017what,conneau2018what,adi2017finegrained,tenney2018what,ettinger2018assessing}. 
Arguments have been made for ``simple'' probes, e.g., that we want to find easily accessible information in a representation \cite{liu2019linguistic,alain2016understanding}.
As a counterpoint though, ``complex'' MLP probes have also been suggested since useful properties might be encoded non-linearly \cite{conneau2018what}, and they tend to report similar trends to simpler probes anyway \cite{belinkov2017what,qian2016investigating}.\looseness=-1

We define control tasks corresponding to English part-of-speech tagging and dependency edge prediction, and use ELMo representations to conduct a broad study of probe families, hyperparameters, and regularization methods, evaluating both linguistic task accuracy and selectivity.
We propose that selectivity be used for building intuition about the expressivity of probes and the properties of models, putting probing accuracies into richer context.
We find that:\looseness=-1
\begin{enumerate}
  \item With popular hyperparameter settings, MLP probes achieve very low selectivity, suggesting caution in interpreting how their results reflect properties of representations. For example, on part-of-speech tagging, $97.3$ accuracy is achieved, compared to $92.8$ control task accuracy, resulting in $4.5$ selectivity.
  \item Linear and bilinear probes achieve relatively high selectivity across a range of hyperparameters. For example, a linear probe on part-of-speech tagging achieves a similar $97.2$ accuracy, and $71.2$ control task accuracy, for $26.0$ selectivity. This suggests that the small accuracy gain of the MLP may be explained by increased probe expressivity.
  \item The most popular method for controlling probe complexity, dropout, does not consistently lead to selective MLP probes. However, control of MLP complexity through unintuitively small (10-dimensional) hidden states, as well as small training sample sizes and weight decay, lead to higher selectivity and similar linguistic task accuracy.
\end{enumerate}

\begin{figure*}
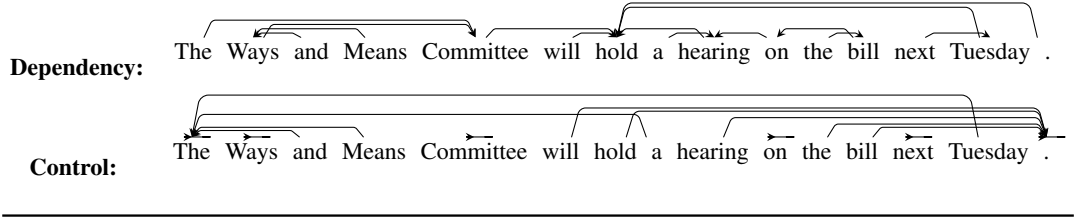

  \centering
  \small

\begin{tabular}{c c}
\multicolumn{2}{c}{\bf Dependency Edge Prediction and Control Task Examples}\\
\midrule
\textbf{Dependency:}  & \begin{dependency}[hide label, edge unit distance=.4ex]
    \begin{deptext}[column sep=0.05cm]
    The\& Ways\& and\& Means\& Committee\& will\& hold\& a\& hearing\& on\& the\& bill\& next\& Tuesday\& . \\
\end{deptext}
\depedge{1}{5}{.}
\depedge{2}{5}{.}
\depedge{3}{2}{.}
\depedge{4}{2}{.}
\depedge{5}{7}{.}
\depedge{6}{7}{.}
\depedge{8}{9}{.}
\depedge{9}{7}{.}
\depedge{10}{9}{.}
\depedge{11}{12}{.}
\depedge{12}{10}{.}
\depedge{13}{14}{.}
\depedge{14}{7}{.}
\depedge{15}{7}{.}
\end{dependency}\\

\textbf{Control: } & \begin{dependency}[hide label, edge unit distance=.3ex]
    \begin{deptext}[column sep=0.05cm]
    The\& Ways\& and\& Means\& Committee\& will\& hold\& a\& hearing\& on\& the\& bill\& next\& Tuesday\& . \\
\end{deptext}
\depedge{1}{1}{.}
\depedge{2}{2}{.}
\depedge{3}{1}{.}
\depedge{4}{1}{.}
\depedge{5}{5}{.}
\depedge{6}{15}{.}
\depedge{7}{15}{.}
\depedge{8}{1}{.}
\depedge{9}{15}{.}
\depedge{10}{10}{.}
\depedge{11}{15}{.}
\depedge{12}{15}{.}
\depedge{13}{13}{.}
\depedge{14}{1}{.}
\depedge{15}{15}{.}
\end{dependency}\\\\
\bottomrule
\end{tabular}
\caption{\label{figuredependencyexample} Example dependency tree from the development set of the Penn Treebank with dependents pointing at heads, and the structure resulting from our dependency edge prediction control task on the same sentence.}
\vspace{-10pt}
\end{figure*}

Finally, we ask, \textit{can we meaningfully compare the linguistic properties of layers of a model using only linguistic task accuracy}?
We raise a potential problem with this approach: it fails to take into account differences in ease of memorization across layers.
In particular, we find that while linear and MLP probes on the first layer of ELMo (ELMo1) achieve slightly higher part-of-speech accuracy than those on the second layer (ELMo2), ($97.2$ compared to $96.6$, for a loss of $0.6$ ), the same probes achieve much greater selectivity on ELMo2 ($31.4$ compared to $26.0$, for a gain of $5.4$).
Thus, the difference in selectivity in favor of ELMo2 is much greater than the commonly known \cite{peters2018deep,liu2019linguistic} difference in linguistic task accuracy in favor of ELMo1; the difference in accuracy may be explained by probes more easily accessing word identity features in ELMo1.

\section{Control Tasks}

In this section, we describe how to construct control tasks.
At a high level, control tasks have:
\begin{description}
  \item[structure:] The output for a word token is a deterministic function of the word type\footnote{Equivalently, word identity.}.
  \item[randomness:] The output for each word type is sampled independently at random.
\end{description}

We start with some notation; denote as $1:T$ the sequence of integers $\{1,...,T\}$.
Let $V$ be the vocabulary containing all word types in a corpus.
A sentence of length $T$ is $\x_{1:T}$, where each $x_i \in V$, and the word representations of the model being probed are $\h_{1:T}$, where $h_i \in \mathbb{R}^d$.
A \textit{task} is a function that maps a sentence to a single output per word, $f(\x_{1:T}) = \y_{1:T}$, where each output is from a finite set of outputs: $y_i \in \mathcal{Y}$.
Each \textit{control task} is defined in reference to a \textit{linguistic task}, and the two share $\mathcal{Y}$. We'll now use part-of-speech tagging and dependency edge prediction as examples to describe the construction of control tasks.

\subsection{Part-of-speech tagging control task}
In part-of-speech tagging, the set $\mathcal{Y}$ is the tagset, $1:45$ (corresponding to NN, NNS, VB,...).
To construct a control task, we independently sample a \textit{control behavior} $C(v)$ for each $v \in V$.
The control behavior specifies how to define $y_i\in\mathcal{Y}$ for a word token $x_i$ with word type $v$.
For part-of-speech tagging, each control behavior directly specifies the output $y_i$ for $x_i$ as an integer from $1:45$, so we sample from 45 behaviors\footnote{The exact distribution from which we sample isn't crucial, but for part-of-speech tagging, we sample from the empirical token distribution of part-of-speech tagging, so the marginal probability of each label is similar.}.
The part-of-speech control task is the function that maps each token $x_i$ to the label specified by the behavior $C(x_i)$:\looseness=-1
\begin{align}
  f_\text{control}(\x_{1:T}) = f(C(x_1),C(x_2),...C(x_T)). \label{eqn_control_task}
\end{align}
This task is visualized in Figure~\ref{figureheaderexplanation}.

\subsection{Dependency edge prediction control task}
The dependency edge prediction task is the function $f_\text{DEP}(\x_{1:T}) = \y_{1:T}$ where $y_i$ is the index of the parent of $x_i$ in the dependency tree on the sentence $\x_{1:T}$.
Thus, the output space $\mathcal{Y} = 1:T$ depends on the length of the sentence, $T$.
To accommodate this in our control task, we define the control behaviors $C(v)$ 
in a length-independent way that still fully specifies $y_i$.
The possible behaviors $C(v)$ are as follows:
\begin{description}
  \item[attach to self:] Always attach tokens of this type to themselves. That is, $y_i = i$.
  \item[attach to first:] Always attach tokens of this type to the first token. That is, $y_i = 1$.\looseness=-1
  \item[attach to last:] Always attach tokens of this type to the last word in the sentence.
    That is, $y_i = T$.
\end{description}
We sample uniformly from the three. Given these behaviors, the control task is defined as before by Eqn~\ref{eqn_control_task}.
This task is visualized in Figure~\ref{figuredependencyexample}.

While very similar to dependency parsing, dependency edge prediction differs in two ways.
The output is not constrained to be a tree for evaluation; each prediction is evaluated independently.
So, while our control tasks do not define trees, the two tasks' output spaces are still the same.
Second, in dependency edge prediction, the root of the sentence is omitted from evaluation; no sentence-external \texttt{ROOT} token is posited for evaluation.

\subsection{Properties of control tasks}
To summarize, a control task is defined for a single linguistic task, and shares the linguistic task's output space $\mathcal{Y}$.
To construct a control task, a control behavior $C(v)$ is sampled independently at random for each word type $v\in V$.
The control task is a function mapping $\x_{1:T}$ to a sequence of outputs $\y_{1:T}$ which is fully specified by the sequence of behaviors, $[C(x_1), ..., C(x_T)]$.

From this construction, we note that the ceiling on performance is the fraction of tokens in the evaluation set whose types occur in the training set (plus chance accuracy on all other tokens.)
  Further, $C(v)$ must be memorized independently for each word type, and a probe taking vectors $h_{1:T}$ as input must identify for each $h_i$ its corresponding $x_i$, and output the element of $\mathcal{Y}$ specified by $C(x_i)$.

\section{Experiments on Probe Selectivity}
In this section, we conduct a broad study of probe families (e.g, linear, MLP) and hyperparameter choices (weight matrix rank/hidden state size, amount of regularization) on a single representation (ELMo1) to determine (1) what probe choices exhibit high linguistic task accuracy \textit{and} high selectivity (and whether this holds for a range of hyperparameters), and (2) whether each probe family can be made selective through hyperparameter choices without substantially sacrificing linguistic task accuracy.

\subsection{Probe families}
We experiment with three types of probes per task.

For part-of-speech tagging, we experiment with linear, MLP-1, and MLP-2 probes. The linear probe is a multiclass model mapping $h_i$ to $y_i \sim \text{softmax}(Ah_i+b)$.
The MLP-1 probe is a multilayer perceptron with one hidden layer and ReLU nonlinearity defined as:
\begin{align}
  &y_i \sim \text{softmax}(W_2\  g(W_1h_i)).
\end{align}
And the MLP-2 probe is defined as:
\begin{align}
  &y_i \sim \text{softmax}(W_3\ g( W_2\  g(W_1h_i))).
\end{align}
where $g$ is the ReLU function, and bias terms are omitted from all affine transformations for brevity.

For dependency edge prediction, we experiment with bilinear, MLP-1, and MLP-2 probes.
These probes take as input the entire sequence $\h_{1:T}$ as well as the vector $h_i$ of a given state to produce $y_i$; the softmax operates over the sequence to construct a distribution over the $T$ classes.
Formally, the bilinear model is defined as $y_i \sim \text{softmax}(\h_{1:T}^\top Ah_i+b)$.
The MLP-1 probe is defined as follows:
\begin{align}
  y_i \sim \text{softmax}(W_2\ g(W_1[\h_{1:T};h_i])).
\end{align}
Note here that $h_i$ broadcasts to $\mathbb{R}^{T\times d}$,while  $W_1 \in \mathbb{R}^{\ell \times d}$, and $W_2 \in \mathbb{R}^{1 \times \ell}$ broadcast as well. That is, each $[h_j;h_i]$ pair is mapped to a single scalar independently of all others, leading to $T$ logits used as input to the softmax.
Similarly, the MLP-2 model is defined as follows:
\begin{align}
  y_i \sim \text{softmax}(W_3\ g(W_2\ g(W_1[\h_{1:T};h_i]))).
\end{align}

\subsection{Complexity control} \label{sectioncomplexitycontrol}
It is well-known that probes should not be too complex \citep{liu2019linguistic,alain2016understanding}; this is the motivation behind constraining the input to the probe to be a single vector or pair of vectors.
However, there has been no systematic investigation of probe complexity.
We study what complexity control is necessary to achieve selectivity.
As we will see, the typical practice of regularizing to reduce the generalization gap (difference between training and test task accuracy) is insufficient if one is interested in selectivity.

\paragraph{Rank/hidden dimensionality constraint.}
For our linear and bilinear probes, we constrain the rank of weight matrices through an LR decomposition.
We let $A\in \mathbb{R}^{k \times d}$, where $k$ is the output space (45 for part-of-speech tagging; 1 for dependency head prediction).
To constrain $A$ to rank $\ell$, we factor $A = LR$, where $L \in \mathbb{R}^{k \times \ell}$ and $R \in \mathbb{R}^{\ell \times |V|}$, and optimize over $L$ and $R$.
For MLP models, we let the hidden state size be equal to $\ell$.\footnote{One could constrain the matrices of the MLP to be rank $\ell$ without making the hidden state smaller, but one must choose a hidden state size anyway, so we believed a study changing the hidden state size would be most informative.}

From the default value of rank-$1000$ and 1000-dimensional hidden states, we let $\ell$ take on the values $\{2,4,10,45\}$ for part-of-speech, and $\{5, 10, 50, 100\}$ for dependency edge prediction.\footnote{Note that for linear models, the rank is constrained by $k$ regardless, since $A\in \mathbb{R}^{k\times d}$.}

\paragraph{Dropout.}
We apply dropout \cite{srivastava2014dropout} with probability $p$ to the input for linear and bilinear probes, and to the input and the output of each hidden layer for MLP probes.
From the default value of $0$, we let $p$ range over $\{0.2, 0.4, 0.6, 0.8\}$.

\paragraph{Number of training examples.}
We artificially constrain the number of sentences the probe is trained on, with the intuition that general rules can be learned more sample-efficiently than memorization.
\citet{zhang2018language} showed this to be an effective distinguishing factor between trained representations and random representation controls.
From the default of $39832$ (the number of training examples in the dataset), we train on $\{4000, 400\}$ examples, corresponding to roughly 100\%, 10\%, and 1\% of the total data, as suggested by \citet{zhang2018language}.

\paragraph{$L_2$ regularization.}
We apply weight decay to the probe parameters.
From the default of $0$, we let the weight decay constant take on the values $\{0.01, 0.1, 1.0, 10.0\}$, unnormalized by batch size.

\paragraph{Early stopping.}
All of our probing models are trained with Adam \cite{kingma2014adam}. By default, we anneal the learning rate by a factor of $0.5$ each time an epoch does not lead to a new minimum loss on the development set, and stop training when 4 such epochs occur in a row.
However, in early stopping, we explicitly halt training at a fixed number of gradient steps.
From the default of $100000$ (approximately 40 epochs), we let this maximum take on the values $\{50000, 25000, 12500, 6000, 3000, 1500\}$.

\begin{figure*}
  \centering
  \includegraphics[width=\linewidth]{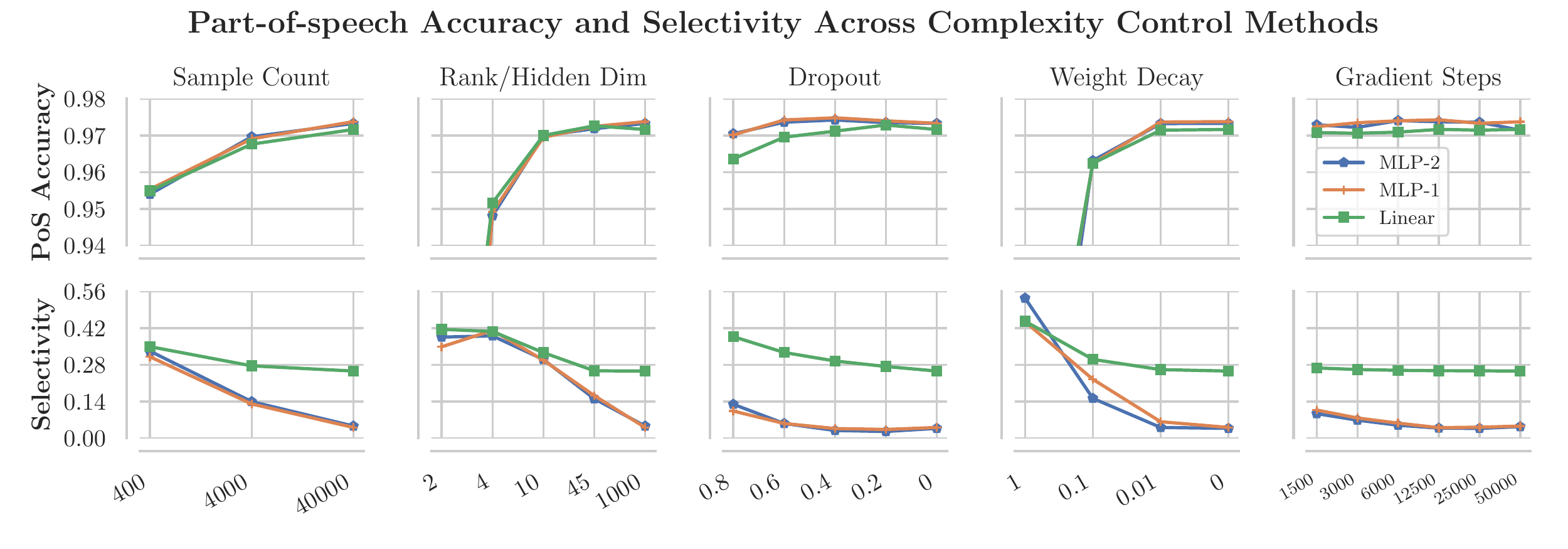}
  \includegraphics[width=\linewidth]{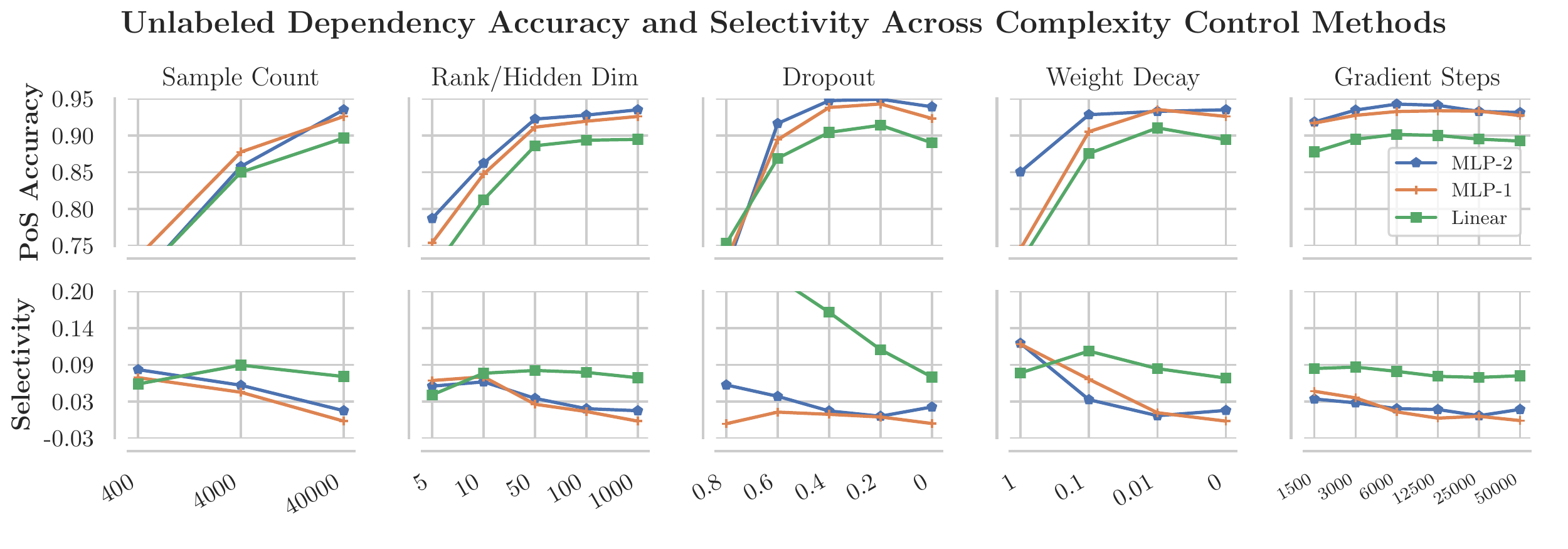}
    \vspace{-25pt}
  \caption{\label{figureallcontrol} Linguistic task accuracies and selectivities for the 5 complexity control methods. All methods except dropout and early stopping are shown to improve selectivity without a large impact on linguistic task accuracy.
    All methods for the same task share a common y-axis, and use their own categorical x-axis.
    All x-axes are ordered from most severe constraints on complexity (left) to most laissez-faire (right).\looseness=-1
    \vspace{-10pt}
  }  
\end{figure*}

\begin{table}
  \centering
  \small
  \begin{tabular}{l  r r r | r r r}
\toprule
\bf Probe & \bf PoS & \bf Ctl & \bf Select. & \bf Dep & \bf Ctl & \bf Select.\\
\midrule
    \multicolumn{7}{c}{Probes with Default Hyperparameters}
\vspace{3pt}\\
Linear & $97.2$ & $71.2$ & $26.0$ & - & - & -\\
Bilinear & - & - & - & $89.0$ & $82.4$ & $6.6$\\
MLP-1 & $97.3$ & $92.8$ & $4.5$ & $92.3$ & $93.0$ & -$0.7$\\
MLP-2 & $97.3$ & $93.2$ & $4.2$ & $93.9$ & $92.0$ & $1.9$\\
\midrule
\multicolumn{7}{c}{Probes with $0.4$ Dropout}
\vspace{3pt}\\
Linear & $97.1$ & $67.3$ & $29.8$ & - & - & -\\
Bilinear & - & - & - & $90.4$ & $73.7$ & $16.7$\\
MLP-1 & $97.5$ & $93.4$ & $4.1$ & $93.8$ & $93.1$ & $0.7$\\
MLP-2 & $97.4$ & $94.1$ & $3.4$ & $94.7$ & $93.5$ & $1.3$\\
\midrule
\multicolumn{7}{c}{Probes Designed with Control Tasks }
\vspace{3pt}\\
Linear & $97.0$ & $64.0$ & $33.0$ & - & - & -\\
Bilinear & - & - & - & $91.0$ & $83.1$ & $7.9$\\
MLP-1 & $97.2$ & $80.6$ & $16.6$ & $90.5$ & $84.3$ & $6.2$\\
MLP-2 & $97.2$ & $81.7$ & $15.4$ & $92.8$ & $89.8$ & $3.0$\\
\bottomrule
\end{tabular}

  \caption{\label{tablesimpleresults} Probe accuracies on linguistic tasks and control tasks.
    Default hyperparameters correspond to a hidden state of dimensionality 1000 and no dropout.
    Under Probes Designed with Control Tasks, we used selectivity to hand-pick a hyperparameter setting for each probe.
    In particular, part-of-speech probes designed with control tasks all use rank-10 weight matrices (10-dimensional hidden state) and no other changes.
  Dependency edge prediction probes designed with control tasks had, for the bilinear model, weight decay of $0.01$, for MLP-1, weight decay of $0.1$, for MLP-2, a rank-50 weight matrix.
  }
\end{table}

\subsection{Dataset}

We use the Penn Treebank (PTB) dataset \cite{marcus1993building} with the traditional parsing training/development/testing splits\footnote{As given by the code of \citet{qi2017arcswift} at \url{https://github.com/qipeng/arc-swift}.} without preprocessing.
We report accuracies on the development set.
We convert the PTB constituency trees to the Stanford Dependencies formalism \cite{marneffe2006generating} for our dependency edge prediction task.

\subsection{Representation}

We use the 5.5 billion-word pre-trained ELMo representations \cite{peters2018deep}.
Since the output of the first BiLSTM layer was recently shown to be the most transferrable on a wide variety of tasks, including part-of-speech and syntax \cite{liu2019linguistic}, we focus on analyzing that layer, which we denote ELMo1.

\subsection{Results} \label{sectionselectivityresults}

\paragraph{Selectivity of default hyperparameters.}
Our results with linear, bilinear, and MLP probes with ``default'' hyperparameters, as specified in \S~\ref{sectioncomplexitycontrol}, are found in Table~\ref{tablesimpleresults} (top).
We find that linear probes achieve similar part-of-speech accuracies to MLPs ($97.2$ compared to $97.3$) with substantially higher selectivity ($26.0$ vs $4.50$).
In dependency edge prediction, we find a definite gap between bilinear probe accuracy ($89.0$) and MLP-1 accuracy ($92.3$).
However, the bilinear probe achieves $16.7$ selectivity, compared to $-0.7$ by MLP-1 and $1.3$ by MLP-2.
Thus, with no regularization, modest gains in linguistic task accuracy through MLP probes over linear/bilinear probes are tempered by losses in selectivity.
Bilinear and linear probes themselves show a significant capacity for memorization.

Does adding moderate regularization through dropout (e.g., $p=0.4$) consistently lead to selectivity? Surprisingly, as shown in Table~\ref{tablesimpleresults} (middle), the opposite is true for some MLP probes, where selectivity actually decreases (e.g., $4.2 \rightarrow 3.4$ for MLP-2).
In one case, the MLP-1 probe on dependency edge prediction, dropout increases selectivity (-$0.7$ $\rightarrow$ $0.7$) but for no others.

\paragraph{How hard is it to find selective probes?}
We tried 6 methods for controlling probe complexity, and all worked except dropout and early stopping, though never for a broad range of hyperparameters.
For each complexity control method except dropout and early stopping, we find hyperparameters that lead to high linguistic task accuracy and high selectivity.
Our results are summarized in Figure~\ref{figureallcontrol}.

We find that constraining the \textbf{hidden state dimensionality} of MLPs is an effective way to encourage selectivity at little cost to linguistic task accuracy.
MLP hidden state sizes of $10$ and $50$, for part-of-speech tagging and dependency head prediction respectively, lead to increased selectivity while maintaining high linguistic task accuracy.
As such, MLP probes with hundreds or $1000$ hidden units, as is common, are overparameterized.

Constraining the \textbf{number of training examples} is effective for part-of-speech, suggesting that learning each linguistic task requires fewer samples than our control task.
However, for dependency edge prediction, this leads to significantly reduced linguistic task accuracy.
Finally, we find that the right \textbf{weight decay} constant can also lead to high-accuracy, high-selectivity probes, especially for dependency edge prediction.
As shown, however, it is unclear what hyperparameters to use (e.g., weight decay $0.1$) to achieve both high accuracy and high selectivity; that is, finding selective MLP probes is non-trivial.

Applying \textbf{dropout}, the most popular probing regularization method \cite{adi2017finegrained,belinkov2019analysis,sahin2019linspector,kim2019probing,elloumi2018analyzing,belinkov2017analyzing,belinkov2018evaluating} does not consistently lead to high-accuracy, high-selectivity MLP probes across a broad range of dropout probabilities ($p=0.2$ to $p=0.8$) on part-of-speech tagging.
For dependency edge prediction, dropout of $p=0.6$ improves the selectivity of MLP-2 but not MLP-1, and considerably increases the already relatively large selectivity of the bilinear probe.
\textbf{Early stopping} in the ranges tested also has little impact on part-of-speech tagging, selectivity, but does improve selectivity of MLP dependency edge prediction probes.\looseness=-1

From our study, we pick a set of hyperparameters for linear, bilinear, MLP-1 and MLP-2 probes to encourage selectivity and linguistic task accuracy together, to compare to default parameters and dropout.
We chose rank constraints of $10$ and $45$, respectively (with no other changes,) for linear and MLP part-of-speech tagging probes, weight decay of $0.01$ for the bilinear dependency probe, and weight decay of $0.1$ for MLP dependency probes.
We report the results of these probes in Table~\ref{tablesimpleresults} (bottom).
In all cases, we see that the right choice of probe leads to considerably higher selectivity than dropout or no regularization.
In particular, for part-of-speech tagging, our chosen MLP-1 probe achieves $16.6$ selectivity, up from $4.5$, and on dependency head prediction, $6.2$ selectivity, up from -$0.7$.\looseness=-1

\subsection{Discussion}

Our most consistent result seems to be that all probes, whether linear, bilinear, or multi-layer perceptron, are over-parameterized and needlessly high-capacity if using defaults like full-rank weight matrices, hidden states with a few hundred dimensions, and moderate dropout.
We can tell this is the case because we're able to heavily constrain the probes (e.g., to rank or 10-dimensional hidden states with little loss in accuracy.

We find that the most selective probes of those tested, even after careful complexity control, are linear or bilinear models.
They also have the advantage that they exhibit high selectivity without the need to search over complexity control methods.

However, the most accurate probes on the more complex task of dependency edge prediction are MLPs, even with hyperparameters tuned for selectivity.
This suggests that while much of the part-of-speech information of ELMo is extractable linearly, some information about syntactic trees is not available to a bilinear function.
In some cases, therefore, one might opt for an MLP probe to extract non-linear features, while optimizing for selectivity through hyperparameter choices.

\paragraph{Errors in Selective and Non-Selective Probes}

Do selective and non-selective probes make different types of errors?
We ran a qualitative study on this, training ten MLP-1 probes and ten linear probes, each with default parameters, on part-of-speech tagging.
We then manually inspected their aggregate confusion matrices for trends in differences between the models' errors.

While the MLP performed marginally better at recognizing many categories, the plurality of improvement over the linear probe by far was in correctly identifying the difference between nouns and adjectives in phrases.
For example,
\begin{quote}
  {\small
  Kan.-based/JJ National/NNP Pizza/NNP\\
  rental/JJ equipment/NN
}
\end{quote}
were correctly labeled by the MLP but not the linear probe, which incorrectly labeled the adjectives as nouns. 
As can be seen with the second example, the distinction between a \textit{JJ NN} modified noun and a \textit{NN NN} noun compound is quite subtle, and the MLP picks up on the distinction considerably better.\looseness=-1

The linear probe, however, was substantially more accurate at predicting the NNP tag, which the MLP probe frequently mislabeled as NNPS.
Manual inspection showed a general trend:
\begin{quote}
  {
  \small
  Environmental/NNP Systems/NNP Co./NNP\\
  Cara/NNP Operations/NNP Co./NNP\\
  7.8/CD \%/NN stake/NN	in/IN	Dataproducts/NNP
}
\end{quote}
In each case, the MLP probe mislabeled the word with the suffix \textit{-s} as NNPS.
The linear probe was considerably less prone to this error.
We hypothesize that this is because the MLP probe is expressive enough to pick up on (spurious) markers  of plurality as well as status as a proper noun independently and combine them, whereas the linear probe is less able to do so.
If this hypothesis is true, then this serves as an example of how less selective probes may be less faithful in representing the linguistic information of the model being probed, since features may be combined to make fine-grained distinctions.

\section{Selectivity Differences Confound Layer Comparisons}

In this section, we use selectivity to shed light on confounding factors when comparing the linguistic capabilities of different representations. 
Multiple studies have found probes on ELMo1 to perform better at part-of-speech tagging than probes on ELMo2 \cite{peters2018deep,tenney2018what,liu2019linguistic}.
As we note, these results depend on the probe as well as the representation; given what we know about probes' capacity for memorizing at the type level, we explore an alternative to the hypothesis that ELMo1 has higher-quality part-of-speech representations than ELMo2.
In particular, word identities are strong features in part-of-speech tagging when used in combination with other indicators; since ELMo1 is closer to the word representations than ELMo2, it may be easier to identify word identities from it, meaning the probe may utilize word identities more readily, as opposed to picking up on a representation of part-of-speech.

\subsection{Experiments}
We run experiments on the first and second contextual layers of ELMo, denoted ELMo1 and ELMo2.
We also examine the representations of an untrained BiLSTM run on the non-contextual character CNN word embeddings of ELMo, shown to be a strong baseline contextualization method, but without any linguistic knowledge learned from context \cite{zhang2018language,hewitt2019structural}. We denote this model Proj0.

We train linear and MLP-1 probes for part-of-speech tagging, and bilinear and MLP-1 probes for dependency edge prediction, all with default hyperparameters (\S~\ref{sectioncomplexitycontrol}).
We examine both the linguistic task accuracy and selectivity achieved by each probe on each representation.

\subsection{Results \& Discussion}
We find probes on ELMo2 to be strikingly more selective than those on ELMo1, consistent across all probes, both for part-of-speech tagging and dependency head prediction.
In particular, the linear probe on ELMo2 achieves selectivity of $31.4$, compared to selectivity of $26.0$ for ELMo1, for a gain of $5.4$.
The same probe achieves $96.6$ linguistic task accuracy on ELMo2 and $97.2$ on ELMo1, for a loss of $0.6$.
The MLP probe shows roughly the same result.
So, does ELMo1 have a better grasp of part-of-speech than ELMo2? 
Our results, summarized in Table~\ref{tablelayerresults}, offer the alternative hypothesis that probes use word identity as a feature to predict part-of-speech, and that feature is less easily available in ELMo2 than ELMo1.

Probes on Proj0 and ELMo2 achieve similar part-of-speech tagging accuracy, echoing findings of \cite{zhang2018language}, but we find that Proj0 is far less selective, suggesting that probes on ELMo2 rely far less on word identities than those on Proj0.
Without considering selectivity, it might be thought that ELMo2 encodes nothing about part-of-speech, since it doesn't beat the Proj0 random representation baseline.
Taking selectivity into account, we see that probes on ELMo2 are unable to rely on word identity features like those on Proj0, so to achieve high accuracy, they must rely on emergent properties of the representation.

\begin{table}
  \centering
  \small
  \begin{tabular}{l r r r r r }
\toprule
\multicolumn{6}{c}{\bf Part-of-speech Tagging} \\ \cmidrule{1-6}
& \multicolumn{2}{c}{Linear} & & \multicolumn{2}{c}{MLP-1} \\ \cmidrule{2-3} \cmidrule{5-6}
\bf Model & Accuracy & Selectivity && Accuracy & Selectivity\\
\midrule
Proj0 & $96.3$ & $20.6$ &  & $97.1$ & $1.6$\\
ELMo1 & $97.2$ & $26.0$ &  & $97.3$ & $4.5$\\
ELMo2 & $96.6$ & $31.4$ &  & $97.0$ & $8.8$\\
\bottomrule
\toprule
\multicolumn{6}{c}{\bf Dependency Edge Prediction} \\ \cmidrule{1-6}
& \multicolumn{2}{c}{Bilinear} & & \multicolumn{2}{c}{MLP-1} \\ \cmidrule{2-3} \cmidrule{5-6}
\bf Model & Accuracy & Selectivity && Accuracy & Selectivity\\
\midrule
Proj0 & $79.9$ & -$4.3$ &  & $86.5$ & -$9.0$\\
ELMo1 & $89.7$ & $6.7$ &  & $92.5$ & -$1.0$\\
ELMo2 & $84.5$ & $6.2$ &  & $89.5$ & $1.4$\\
\bottomrule
\end{tabular}
   \caption{\label{tablelayerresults} Part-of-speech and dependency edge prediction probe accuracies and selectivities across three representations. ELMo1 and ELMo2 are the two contextual layers of ELMo, while Proj0 refers to an untrained BiLSTM contextualization of ELMo's non-contextual character CNN representations.}
\end{table}

\section{Related Work}

Early work in probing, (also known as diagnostic classification \cite{hupkes2018visualisation},) extracted properties like parts-of-speech, gender, tense, and number from distributional word vector spaces like word2vec and GloVe \cite{mikolov2013distributed,pennington2014glove} using linear classifiers \cite{kohn2015s,gupta2015distributional}.
Soon after, the investigation of intermediate layers of deep models using linear probes was introduced independently by \citet{ettinger2016probing} and \citet{shi2016does} in NLP and \citet{alain2016understanding} in computer vision.

Since then, probing methods have varied as to whether they investigate whole-sentence properties like sentence length and word content using a sentence vector \cite{shi2016does,adi2017finegrained,conneau2018what}, word properties like verb tense or part-of-speech using word vectors \cite{shi2016does,belinkov2017what,liu2019linguistic}, or word-pair properties like syntactic relationships using pairs of vectors \cite{tenney2018what,hewitt2019structural}.
Probes have been used to make relative claims between models or components \cite{adi2017finegrained,liu2019linguistic,belinkov2017what} or absolute claims about models above baselines.
Probes have also been used to test hypotheses about the mechanisms by which models perform tasks \cite{hupkes2018visualisation,giulianelli2018hood}.

Previous work has made extensive use of control representations like non-contextual word embeddings or models with random weights \cite{belinkov2017what,tenney2018what,saphra2018understanding,hewitt2019structural}; our control tasks provide a complementary perspective, measuring a probe's ability to decode a random function from the representation of interest.

The most related work to ours is that of \citet{zhang2018language}, who presented experiments for understanding the roles probe training sample size and memorization have on linguistic task accuracy.
They observed that untrained BiLSTM contextualizers achieved almost the same part-of-speech tagging accuracies as trained contextualizers, and found that by reducing the probe training set, the trained models could be shown to significantly outperform the untrained model.
They evaluated which representations were easiest to memorize from by probing to predict nearby words, finding as we do that word identities are most easily available in untrained contextualizers' representations.
They take this as evidence that gains in part-of-speech probing accuracy on the trained representations over the untrained representations are due to linguistic properties, not memorization.
Our experiments with selectivity complement their results, finding among other things that even though untrained BiLSTMs are better for memorization than ELMo, there is still a striking capacity for memorization using ELMo when using high-capacity probes. \looseness=-1

\subsection{Random tasks}
\citet{zhang2018understanding} defined completely random tasks related to Rademacher complexity \cite{bartlett2001rademacher} to understand the capacity of neural networks to overfit, showing that they are expressive enough to fit random noise, but still function as effective models.
In our random control tasks, randomness is applied at the type-level rather than at the example-level, and are designed to have strong non-linguistic structure as opposed to absolutely no structure.
While the tasks of \citet{zhang2018understanding} aid in understanding the expressivity of neural nets, our control tasks aid in understanding the expressivity of a probe model with respect to a specific linguistic task.

\section{Conclusion}
Through probing methods, it has been shown that a broad range of supervised learning tasks can be turned into tools for understanding the properties of contextual word representations \cite{conneau2018what,tenney2018what}.
\citet{alain2016understanding} suggested we may think of probes as ``thermometers used to measure the temperature simultaneously at many different locations''.
We instead emphasize the joint roles of representations and probes together in achieving high accuracy on a task; we suggest that probes be thought of as \textit{craftspeople}; their performance depends not only on the materials they're given, but also on their expressivity.

To explore the relationship between representations, probes, and task accuracies, we defined control tasks, which by construction can only be learned by the probe itself.
We've suggested that a probe which provides insights into the properties of the representation should be \textit{selective}, achieving high linguistic task accuracy and low  control task accuracy.
Selectivity measures the probe's ability to make numerous output decisions independently of linguistic properties of the representation.

We've found that linear and bilinear models achieve higher selectivity at similar accuracy to MLP probes on part-of-speech tagging.
MLP probes, achieving higher accuracy on the more complex task of dependency edge prediction, can be re-designed to achieve higher selectivity at a relatively small cost to dependency edge accuracy, but often not through dropout, the most popular MLP probe regularization method.

Finally, we showed how selectivity can be used to provide added context to probing results, demonstrating that marginal differences in part-of-speech tagging accuracy between ELMo1 and ELMo2 correspond to large differences in selectivity, and similarly, the even though ELMo2 achieves similar part-of-speech tagging accuracy to a random representation baseline, ELMo2 achieves it with much higher selectivity.

As probes are used increasingly to study representations, we hope that control tasks and selectivity, as diagnostic tools, can help us better interpret the results of these probes, ultimately leading us to better understand what is learned by these remarkably effective representations.

\paragraph{Reproducibility.} All code, data, and experiments are available at \url{https://worksheets.codalab.org/worksheets/0xb0c351d6f1ac4c51b54f1023786bf6b2}.

\section*{Acknowledgements}
We would like to thank Kevin Clark, Robin Jia, Peng Qi, Vivek Kulkarni, Surya Ganguli, Tatsu Hashimoto, and Nelson Liu for helpful discussions and feedback.
We would like to thank our reviewers for clarifying comments and suggestions for extra experiments.
This work was funded under a PECASE Award.

\bibliography{emnlp-ijcnlp-2019}
\bibliographystyle{acl_natbib}

\end{document}